\documentclass{article} 
\usepackage{iclr2015,times}
\usepackage{hyperref}
\usepackage{url}
\usepackage{graphicx}
\usepackage{upgreek}
\usepackage{amssymb}
\usepackage{algorithm}
\usepackage{algpseudocode}
\usepackage{bm}
\usepackage{dsfont}
\usepackage{tabularx}
\usepackage{multirow}
\usepackage{longtable}
\usepackage{footnote}
\usepackage{standalone}
\usepackage{booktabs, dcolumn}
\title{Training Binary Multilayer Neural Networks for Image Classification using Expectation Backpropagation}

\author{
Zhiyong Cheng \\
School of Information Systems\\
Singapore Management University\\
Singapore 178902, Singapore \\
\texttt{zy.cheng.2011@smu.edu.sg} \\
\And
Daniel Soudry \\
Department of Statistics\\
Columbia University\\
\texttt{daniel.soudry@gmail.com} \\
\And
Zexi Mao \& Zhenzhong Lan \\
School of Computer Science\\
Carnegie Mellon University\\
Pittsburgh, PA 15213, USA \\
\texttt{\{zexim, lanzhzh\}@cs.cmu.edu} \\
}

\begin{document}

\maketitle

\begin{abstract}
Compared to Multilayer Neural Networks with real weights, Binary Multilayer Neural Networks (BMNNs) can be implemented more efficiently on dedicated hardware. BMNNs have been demonstrated to be effective on binary classification tasks with Expectation BackPropagation (EBP) algorithm on high dimensional text datasets. In this paper, we investigate the capability of BMNNs using the EBP algorithm on multiclass image classification tasks.  The performances of binary neural networks with multiple hidden layers and different numbers of hidden units are examined on MNIST. We also explore the effectiveness of image spatial filters and the dropout technique in BMNNs. Experimental results on MNIST dataset show that EBP can obtain 2.12\% test error with binary weights and 1.66\% test error with real weights, which is comparable to the results of standard BackPropagation algorithm on fully connected MNNs.
\end{abstract}

\section{Introduction}
In recent years, deep neural networks (DNNs) have attracts tremendous attentions from a wide range of research areas related to signal and information processing.  State-of-the-art performances have been achieved with DNN techniques on various challenging tasks and applications, such as speech recognition~\citep{hinton2012deep}, object recognition~\citep{Krizhevsky2012,szegedy2014going},  multimedia event detection~\citep{lan2013cmu}, etc. Almost all the current DNNs are real-valued-weight Mutlilayer Neural Networks (RMNNs). However, an effective RMNNs are often massive and require large computational and energetic resources. For example, GoogLeNet has 22 layers with tens of thousands of hidden units~\citep{szegedy2014going}. MNNs with binary weights (BMNNs) have the advantage that they can be implemented efficiently on dedicated hardware. For example, \citet{karakiewicz2012} have presented a chip which enable  $10^{12}$ multiply accumulates per second per mW power efficiency with binary weights. Thus, it is attractive to develop effective algorithms for BMNNs to achieve
comparable performances with RMNNs.

Traditional MNNs are trained with BackPropagation (BP) or similar gradient descent methods. However, BP or gradient descent methods cannot
be directly used for training binary neural networks.  A straightforward method for this problem is to binarize the real-valued weights, while this approach will decrease the performance significantly.
Recently, ~\citet{Soudry14} presented an Expectation BackPropagation (EBP) algorithm, which can support online training of MNNs with either continuous or discrete weight values. Experiments on several large text datasets show promising performances on binary classification tasks with binary-weighted MNNs~\citep{Soudry14}.  As an extension of the previous work by ~\citet{Soudry14},
in this work, we study the performance of EBP algorithm on image classification tasks with binary and real weights MNNs.
Besides, we investigate the effects of different factors,  such as network depth, layer size and dropout strategies,
on the performance of EBP algorithm in image classification. This study
explores the possibility of using BMNNs for the multimedia supervised classification tasks.

\section{Expectation Backpropagation}
In this section, we review the expectation backpropagation (EBP) and  introduce how to implement the EBP algorithm for binary weights in detail.
Before introducing the EBP algorithm, we first describe the general notations.

A blodfaced capital letter $\bm{X}$ denotes a matrix with components $X_{ij}$. A blodfaced non-capital letter $\textbf{x}$ denotes a column vector with components $x_i$. Besides, $\textbf{x}_l$ denotes $x_{i,l}$ and $\bm{X}_l$ denotes $X_{ij,l}$. The indicator function $\mathcal{I}(A)$ denotes that $\mathcal{I}(A)=1$ if condition $A$ holds, and 0 otherwise. We consider a general feedforward Multilayer Neural Networks (MNN) with connections only between adjacent layers. Suppose the MNN has $L$ layers, $V_l$ is the number of hidden units in the $l$-th layer, and $\mathcal{W} = \{\bm{W}_{l}\}^L_{l=1}$ is weight matrices $V_l \times V_{l-1}$ between the $(l-1)$-th layer and $l$-th layer. For simplicity, the activation function is $\bm{v}_l=sign(\bm{W}_l\bm{v}_{l-1})$ function in this study. The output of the network is therefore
\begin{equation} \label{eq:ebp-d}
    \textbf{v}_L = g(\textbf{v}_0, \mathcal{W})=sign(\bm{W}_Lsign(\bm{W}_{L-1})sign(...\bm{W}_1\textbf{v}_0))
\end{equation}\

Similar to supervised learning with MNNs, the task is to learn $\mathcal{W}$ for a MNN with known architecture given a set of labeled data pairs $D_N = \{\textbf{x}^{(n)}, \textbf{y}^{(n)}\}^N_{n=1}$ (note $D_0 = \emptyset$), where each $\textbf{x}^{(n)} \in \mathds{R}^{V_0}$ is a data point, and each $\textbf{y}^{(n)} \in \{-1,+1\}^{V_L}$ is a label.

The EBP algorithm is derived within Bayesian framework. Given the labeled dataset, the aim is to find the weights $\mathcal{W}$ to maximize the posterior probability $P(\mathcal{W}|D_N)$. With the posterior, one can obtain the most probable weight configuration to minimize the expected zero-one loss over the outputs using the Maximum A Posteriori (MAP) estimation.
\begin{equation} \label{eq:ebp-p}
    \textbf{y}^* = argmax_{\textbf{y} \in \mathcal{Y}}\sum_{\mathcal{W}}\mathcal{I}\{g(\textbf{x}, \mathcal{W})=\textbf{y}\}P(\mathcal{W}|D_N)
\end{equation}
The posterior $P(\mathcal{W}|D_N)$ is updated in an online setting, where samples arrive sequentially. According to the Bayes rule, when the $n$-th sample is arrived, the posterior is updated as follows. For $n=1, ..., N$,
\begin{equation}
    P(\mathcal{W}|D_n) \propto P(\textbf{y}^{(n)}|\textbf{x}^{(n)}, \mathcal{W})P(\mathcal{W}|D_{n-1})
\end{equation}
However, this update is generally intractable for large networks, as there is an exponential number of values for $P(\mathcal{W}|D_{n})$ to be stored and updated. To solve this problem, the mean-field approximation is used to approximate $P(\mathcal{W}|D_{n})$. Specifically,  $P(\mathcal{W}|D_{n})$ is approximated by $\hat{P}(\mathcal{W}|D_n)$, for which
\begin{equation}
    \hat{P}(\mathcal{W}|D_n) = \prod_{i,j,l}\hat{P}(W_{ij,l}|D_n)
\end{equation}
where each factor is normalized. Based on the equation, performing a marginal of the posterior (see appendix A in~\citet{Soudry14} for details) of the Bayes update and re-arrange terms, we can obtain a Bayes-like update to the marginal
\begin{equation} \label{eq:posup}
\hat{P}(\mathcal{W}|D_n) \propto \hat{P}(\textbf{y}^{(n)}|\textbf{x}^{(n)}, W_{ij,l}, D_{n-1})\hat{P}(W_{ij,l}|D_{n-1})
\end{equation}
where
\begin{equation} \label{eq:sigleupdate}
   \hat{P}(\textbf{y}^{(n)}|\textbf{x}^{(n)}, W_{ij,l}, D_{n-1}) = \sum_{\mathcal{W}':W'_{ij,l}=W_{ij,l}}P(\textbf{y}^{(n)}|\textbf{x}^{(n)},\mathcal{W})\prod_{\{k,r,m\} \neq \{i,j,l\}}\hat{P}(W'_{kr,m}|D_{n-1})
\end{equation}
is the marginal likelihood. Accordingly, the $\hat{P}(\mathcal{W}|D_n)$ can be directly updated in a single step. The problem is that Eq.~\ref{eq:sigleupdate} contains a generally intractable summation over an exponential number of values.

To simplify the summation, another approximation is performed by assuming that the neuronal fan-in is ``large'', namely, a large number of units in the previous layer is connected to each unit in the next layer. Since all the other weights besides $W_{ij,l}$ are independent (based on the mean field approximation), together with the large fan-in assumption, we can assume that the normalized input to each neural layer is a Gaussian distribution based on the Central Limit Theorem (CLT), thus
\begin{equation} \label{eq:gaussian}
 \forall{m}: \textbf{u}_m=\textbf{W}_m\textbf{v}_{m-1}/\sqrt{K_m} \sim \mathcal{N}(\bm{\mu}_m, \bm{\Sigma}_m)
\end{equation}
This is a quite common and effective one~\citep{ribeiro2011expectation} approximation. Using this approximation (Eq.~\ref{eq:gaussian}) and the activation function $\textbf{v}_m=sign(\textbf{u}_m)$, the distribution of $\textbf{u}_m$ and $\textbf{v}_m$ can be calculated sequentially for all the layers $m \in \{1,...,L\}$ (``\textbf{forward pass}"), for any given value of $\textbf{v}_0$ (i.e., the input) and $W_{ij,l}$. At the end of the forward pass, we can obtain $P(\textbf{y}|W_{ij,l})=P(\textbf{v}_L =\textbf{y}|W_{ij,l})$, $\forall {i,j,l}$. With the obtained $P(\textbf{y}|W_{ij,l})$, we can use Eq.~\ref{eq:posup} to update $W_{ij,l}$, $\forall {i,j,l}$.

Because it is very computational to directly calculate $P(\textbf{v}_L =\textbf{y}|W_{ij,l})$ for every $i, j, l$, Taylor expansion of $W_{ij,l}$ (around its mean, $\langle W_{ij,l} \rangle$ to first order) is used to approximate $P(\textbf{v}_L =\textbf{y}|W_{ij,l})$. The first order terms in this expansion can be calculated using \textbf{backward propagation} of derivative terms
\begin{equation} \label{eq:deriv}
    \Delta_{k,m}=\partial{\ln{P(\textbf{v}_L=y)}}/\partial{\mu_{k,m}}
\end{equation}
Thus, after a forward pass for $\textbf{u}_m$ and $\textbf{v}_m$, $m \in \{1,...,L\}$, and a backward pass for $P(\textbf{v}_L =\textbf{y}|W_{ij,l})$, $l \in \{L,...,1\}$ for all $W_{ij,l}$, we can update $P(W_{ij,l})$ in each training epoch. In the next, we will summarize the general Expectation BackPropagation algorithm and introduce the implementation of EBP algorithm using binary weights and real bias. More detailed information about the implementation for real weights is described in \citet{Soudry14}.

\subsection{The Expectation Backpropagation Algorithm}
Given input $\textbf{x}$ and desire output $\textbf{y}$, a forward pass is first performed to calculate the mean output $\langle v_l \rangle$ for each layer; then a backward pass is conducted to update $P(W_{ij,l}|D_n)$ for all the weights.

\textbf{Forward pass} First, we initialize the MNN input $\langle v_{k,0} \rangle=x_k$ for all $k$, and then calculate recursively the following values for $m = 1, ..., L$ and all $k$
\begin{equation} \label{eq:forward1}
    \mu_{k,m} = \frac{1}{\sqrt{K_m}}\sum^{V_{m-1}}_{r=1}\langle W_{kr,m} \rangle\langle v_{r,m-1}\rangle; \hspace{0.3cm} \langle v_{k,m} \rangle = 2 \upphi (\mu_{k,m}/\sigma_{k,m})-1
\end{equation}

\begin{equation}\label{eq:forward2}
    \sigma^2_{k,m}=\frac{1}{K_m}\sum^{V_{m-1}}_{r=1}\langle W^2_{kr,m} \rangle(\delta_{m,1}(\langle v_{r,m-1} \rangle^2 - 1)+1)-\langle W_{kr,m} \rangle^2\langle v_{r,m-1} \rangle^2
\end{equation}

where $\langle W_{kr,m} \rangle$ is the mean of the posterior distribution $P(W_{ij,l}|D_n)$. $\mu_m$ and $\sigma^2_m$ are the mean and variance of $u_m$ of the input of layer $m$, and $\langle v_m \rangle$ is the resulting mean of the output of layer $m$

\textbf{Backward pass} The backward pass performs the Bayes update of the posterior (Eq.~\ref{eq:posup}) using a Taylor expansion. Based on Eq.~\ref{eq:deriv}, we first initialize $\Delta_{i,L}$ for all $i$ (refer to the Eq. C.9 in~\citep{Soudry14}) as:
\begin{equation}
    \Delta_{i,L} = y_i\frac{\mathcal{N}(0|\mu_{i,L},\sigma^2_{i,L})}{\upphi(y_i\mu_{i,L}/\sigma_{i,L})}
\end{equation}
Then, for $l=L, ..., 1$ and $\forall{i,j}$, we calculate
\begin{equation}
    \Delta_{i,l-1} = \frac{2}{\sqrt{K_l}}\mathcal{N}(0|\mu_{i,l-1},\sigma^2_{i,l-1})\sum^{V_m}_{j=1}\langle W_{ji,l} \rangle\Delta_{j,l}
\end{equation}
\begin{equation} \label{eq:backward2}
    \ln{P(W_{ij,l})|D_n}=\ln{P(W_{ij,l}|D_{n-1})}+\frac{1}{\sqrt{K_l}}W_{ij,l}\Delta_{i,l}\langle v_{j,l-1} \rangle + C
\end{equation}
where C is an unimportant constant, which is not dependent on $W_{ij,l}$.

\textbf{Output} Based on the learnt weight configuration $\mathcal{W}^*$, the output can be obtained by $g(\bm{x}, \mathcal{W})$ by Eq.~\ref{eq:ebp-d}, which is defined as the Deterministic EBP output (\textbf{EBP-D})~\citep{Soudry14}. Alternatively, the MAP output (Eq.~\ref{eq:ebp-p}) can be calculated directly
\begin{equation} \label{eq:ebpp}
    \textbf{y}^* = argmax_{y \in \mathcal{Y}}\ln P(\textbf{v}_L=\textbf{y})=argmax_{y \in \mathcal{Y}}[\sum_k\ln(\frac{1+\langle v_{k,L} \rangle}{1 - \langle v_{k,L} \rangle})^{y_k}]
\end{equation}
using $\langle v_{k, L} \rangle$ from Eq.~\ref{eq:forward1}. The output of Eq.~\ref{eq:ebpp} is defined to be the Probabilistic EBP output (\textbf{EBP-P}).

\subsection{Implementation for Binary Weights}
In the implementation of binary weights, the weight $w_{ij,l}$ can only take value $\{-1,+1\}$. In \citet{Soudry14}, the distribution of $W_{ij,l}$ is parameterized in the way so that
\begin{equation} \label{eq:bin}
    P(W_{ij,l}|D_n) = \frac{e^{h^{(n)}_{ij,l}W_{ij,l}}}{e^{h^{(n)}_{ij,l}} + e^{-h^{(n)}_{ij,l}}}
\end{equation}

According to the forward process (Eq.~\ref{eq:forward1} and Eq.~\ref{eq:forward2}), the parametrization can be used to compute $\langle W_{ij,l} \rangle = tanh(h_{ij,l})$, $\langle W^2_{ij,l} \rangle = 1$ and $Var(W_{ij,l})=sech^2(h_ij,l)$. In the backward processing, substituting Eq.~\ref{eq:bin} into Eq.~\ref{eq:backward2}, the parameter $h^{(n)}_{ij,l}$ is updated in each iteration as
\begin{equation}
    h^{(n)}_{ij,l} = h^{(n-1)}_{ij,l} + \frac{1}{\sqrt{K_l}}{\Delta_{i,l}}{\langle v_{j,l-1} \rangle}
\end{equation}

\textbf{Algorithm 1} shows the update steps of the EBP algorithm for BMNN.  The weight configuration for the BMNN is obtained by simply clipping
\begin{equation}
   W^*_{ij,l}=sign(h_{ij,l})
\end{equation}

\begin{algorithm}
\caption{Expectation BackPropagation (EBP) algorithm for fully connected binary MNNs - with binary synaptic weights and real bias~\citep{Soudry14}}.
\label{algorithm1}
\begin{algorithmic}
\State \% $\nu_{k,l}=\langle v_{k,l} \rangle$, $tanh(h_{ij,l}) = \langle W_{ij,l} \rangle$, and $\mathcal{H}$ is the set of all $h_{ij,l}$.
\State \textbf{Function} $[\bm{\nu}_L$, $\bm{\mathcal{H}}_{next}] = $ UpdateStepBinaryMNN($\textbf{x}$, $\textbf{y}$, $\bm{\mathcal{H}}$) \newline
\Comment{\textbf{Forward pass}}
\State \textbf{Initialize:}
    \begin{center}
        $\forall{k}: \nu_{k,0}=x_k, \forall{l}: \nu_{0,l}=1$
    \end{center}
\For {$m=1 \to L$} \newline
    $\forall{k}:$

            $\mu_{k,m} = \frac{1}{\sqrt{K_{m-1}}}[h_{k0,m} + \sum^{V_{m-1}}_{r=1}tanh(h_{kr,m})\nu_{r,m-1}]$

            $\sigma^2_{k,m} = \frac{1}{K_{m-1}}[1+\sum^{V_{m-1}}_{r=1}[(1-\nu^2_{r,m-1})(1-\delta_{1m}) + \nu^2_{r,m-1}sech^2(h_{kr,m})]]$

            $\nu_{k,m}=2\upphi(\mu_{k,m}/\sigma_{k,m})-1$ \newline

\EndFor \newline
\Comment{\textbf{Backward pass}}
\State \textbf{Initialize:}
    \begin{center}
        $\Delta_{i,L}=y_i\frac{\mathcal{N}(0|\mu_{i,L},\sigma^2_{i,L})}{\upphi(y_i\mu_{i,L})/\sigma_{i,L}}$
    \end{center}
\For {$l=L \to 1$} \newline

        $\forall{i}: \Delta_{i, l-1} = \frac{1}{\sqrt{K_{l-1}}}\mathcal{N}(0|\mu_{i,l-1},\sigma^2_{i,l-1})\sum^{V_{m}}_{j=1}tanh(h_{ji,l})\Delta_{j,l}$

        $\forall{i,j}: h^{next}_{ij,l}=h_{ij,l}+ \frac{1}{\sqrt{K_{l-1}}}\Delta_{i,l}\mu_{j,l-1}$ \newline

\EndFor

\end{algorithmic}
\end{algorithm}

\section{Implementation of EBP on Image Classification}
The performance of EBP algorithm has been evaluated in~\citet{Soudry14}. However, those experiments are limited to high dimensional text datasets (the dimensions of the input feature vectors are from 11,463 to 238,739), and all the tasks are binary classification tasks. In this study, we will examine the performance of the EBP algorithm on image datasets for multiclass classification. To check the performance of EBP algorithm on deeper and small ``fan-in" architectures on  image classification,  we use architectures with multiple layers and different hidden unites in experiments. Besides, we also explore the effectiveness of dropout techniques~\citep{srivastava2014dropout} in EBP algorithms.

\begin{figure}[h]
\begin{center}
\includegraphics[width= 7cm]{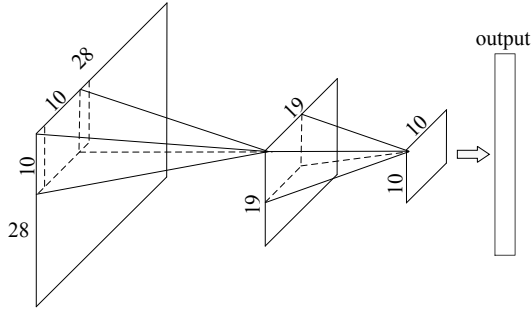}
\end{center}
\caption{A two-layer neural network architecture that considers the spatial context in images}
\label{fig:2d}
\end{figure}

Two methods are used to input the image into the MNNs. The first method is to directly convert the 2D image into 1D vector by concatenating the pixels in the image in certain order, such as concatenating each row from top to bottom. For example, for the standard MNIST handwritten digits database, the input of each image is $28 \times 28$ vector. In the second method (spatial filtering method), we consider the spatial configuration of the images. The spatial configuration is considered in a similar way as Convolutional Neural Networks (CNN)~\citep{lecun1998gradient}. Each unit in a layer receives inputs from a set of units located in a small neighborhood in the previous layer.  As shown in Fig.~\ref{fig:2d}, a unit in the feature map has 100 inputs connected to a $10 \times 10$ area in the input. Each unit has 100 inputs and therefore 100 trainable coefficients plus a trainable bias. Different from CNN, we only use one feature map in each hidden layer in this study.\footnote{The performance of EBP algorithm on standard CNN architectures will be studied in further work.} Since there is only one feature map, the network does not have the constraint that the connection weights for each unit in the feature map are the same. In the example shown in Fig.~\ref{fig:2d}, there are $19 \times 19 = 361$ units in the second layer and each unit have (100 + 1) trainable weights. In implementation, the weight matrix between the first and second layer is set to $361 \times 784$. The weight matrix is initialized in the way that only the weights for connected units are nonzero, namely, $361 \times 100$ nonzero elements in the weight matrix. And the zero elements are kept zero during the whole training process. Because the EBP algorithm have the assumption of large fan-in, each unit in the hidden layers (feature maps) should be connected to a relative large neighborhood (such as ``$10 \times 10$" or larger) in the input layer.

\section{Experiments}
In this section, we report the experiments of the EBP algorithm with MNNs with different architecture configurations in the standard MNIST handwritten digits database~\citep{lecun1998gradient}.

\subsection{Experiment Setup}
The MNIST database contains 60,000 images ($28 \times 28$ pixels) and the test set has other 10,000 images. During the training process, all the images in the training set were presented sequentially in each epoch with a randomized order. The task was to identify the label $\{0,1,...,9\}$, using a BMNN classifier trained by EBP algorithm. The label is set to be $y_k = 2\delta_{k,label+1} + 1$. We pre-process the training data by centralizing ($mean=0$) and normalizing ($std=1$) the pixels as recommended for BackPropagation~\citep{lecun2012efficient}.  As standard for classification with real values of MNNs. The output neuron with highest value indicates predicted label of the input pattern.

When treating the image as 1D vector, a constant 1 is added to each input vector to allow some bias to the neurons in the hidden layer (so $\textbf{v}_0 = 785$). For the spatial filtering method, a bias is added to each convolving block.  Two neural network architectures are used: one hidden layer and two hidden layers.  For each type of architecture, we vary the number of neurons in the hidden layers. The detailed configurations for the network architectures for both methods are shown in Table~\ref{tab:config}. In the spatial filtering method, different filtering block sizes are used in the one hidden layer architecture: $12 \times 12$, $13 \times 13$, $14 \times 14$, $15 \times 15$, $16 \times 16$ and $17 \times 17$. Thus, the corresponding hidden units are 289, 256, 255, 196, 169 and 144, which are the feature map size in the hidden layer. Taking block size $12 \times 12$ as an example, the feature map size becomes $(28 - 12 + 1) \times (28 - 12 + 1) = 17 \times 17 = 289$ hidden units. Accordingly, there are $12 \times 12 = 144$ inputs to each unit in the hidden layer, and 289 inputs to each unit in the output layer. We selected such configurations because of the large ``fan-in" assumption of the EBP algorithm. These configurations can also be used to learn whether it is better to set larger fan-in in the first layer or second layer. In the case of two-hidden-layer network, we only select one configuration because other configuration will lead to smaller fan-in (the hidden units [361, 100] correspond to 100 inputs to each unit in the first layer ($10 \times 10$ block size in the input layer) and also 100 inputs ($10 \times 10$ block size in the second layer) to each unit in the second hidden layer).

\begin{table}
\centering
\caption{Network architectures in experiments}
\begin{tabular}{|l||c||l|}
 \hline
    Method  & \# of Hidden Layers & \# of Hidden Units \\ \hline \hline
    \multirow{2}{*}{1D Vector}  & One & 200; 400; 600; 800; 1000 \\ \cline{2-3}
     & Two & [200, 200]; [400, 400]; [600, 600]; [800, 800] \\ \hline
    \multirow{2}{*}{2D Convolving} & One & 144; 169; 196; 255; 256; 289 \\ \cline{2-3}
     & Two & [361, 100] \\\hline

\end{tabular}
 \label{tab:config}
\end{table}

We also employ dropout technique on all the architectures. Dropout is a technique for preventing overfitting and provides a way of approximately combining exponentially many different network architectures efficiently to improve performance~\citep{srivastava2014dropout}. The effectiveness of dropout has been demonstrated on neural networks, DBN and DBM with traditional error backpropagation with stochastic gradient decent method~\citep{srivastava2014dropout}. In this study, we investigate its effectiveness in the EBP algorithm. In the experiments, we fixed $p=0.8$  for both hidden units and input units in all dropout nets.

\subsection{Experimental Results}
In the result presentation, we use four abbreviations for presentation simplicity: (1) B-EBP-D:  Deterministic EBP (EBP-D, see Sect. 2.1) with binary weights;  (2) B-EBP-P: Probabilistic EBP (EBP-P, see Sect. 2.1) with binary weights; (3) R-EBP-D: Deterministic EBP with real weights; and (4) R-EBP-P: Probabilistic EBP with real weights. All the results reported below are based on the networks trained by 120 epochs.  Training with more epochs may improve the performances of some network architectures. For weight initialization, we used the same method as~\citet{Soudry14}.

\textbf{Effects of Hidden Unite Number and Hidden Layer Number} Table~\ref{tab:1d_without_dropout} shows the results of MNNs on MNIST dataset using EBP algorithms on different network structures without dropout. From the results, we can observe that for networks with one hidden layer, the increase of hidden units clearly improves the performance and the best performance is obtained with 800 units. Two-hidden-layer structure with EBP-P outperforms the one-hidden-layer structure significantly, even with only 200 hidden units in each layer. The results demonstrate the EBP works well on MNNs. Another observation is that EBP-P outperforms EBP-D, which is consistent with the results shown in~\citet{Soudry14}. Particularly, the performance of B-EBP-D in the two-hidden-layer structure is worse than that of in the one-hidden-layer structure. With growing size of hidden units, performance of B-EBP-D decreases quickly in two-hidden-layer models. We also use the EBP algorithm with real weights for all the configurations. The performances of EBP with real weights are better than the performance of EBP with binary weights in all structures. R-EBP-P in two-hidden-layer is only slightly better than  in one-hidden-layer. Although R-EBP-D in two-hidden-layer performs worse than in one-hidden-layer as B-EBP-D, its performance increase  when the number of hidden units increases. The standard BackProp algorithm (using $tanh$ activation function and optimized learning rate) on the one-hidden-layer model with 800 units can obtain 2.13\%\footnote{Note that using error regularization and proper weight initialization, standard backpropagation can achieve better performance. For example, we can achieve 1.65\% error rate by using L1 and L2 error regularization and initializing the weight uniformly in $[-\sqrt{\frac{6}{fan_{in}+fan_{out}}}, \sqrt{\frac{6}{fan_{in}+fan_{out}}}]$ with 500 hidden units.}, which is comparable for the best results obtained by R-EBP-P. Using binary weight will hurt the performance, while from the table, we can see that binary weights with optimal neural networks do not hurt the performance much (best performance of B-EBP-P is 2.37\%, comparing to 2.10\% of R-EBP-P).

\begin{table}
\scriptsize
\centering
\caption{Test errors without dropout for 1D vector method}
\begin{tabular}{|c||c|c|c|c|c||c|c|c|c|} \hline
Hidden units	&	200	&	400	&	600	&	800	&	1000	&	[200, 200]	&	[400, 400]	&	[600, 600]	&	[800, 800]	\\ \hline \hline
B-EBP-P	&	3.46\%	&	3.15\%	&	3.12\%	&	3.01\%	&	3.11\%	&	2.63\%	&	2.61\%	&	2.37\%	&	2.37\%	\\ \hline
B-EBP-D	&	4.63\%	&	3.89\%	&	3.62\%	&	3.63\%	&	3.57\%	&	5.20\%	&	5.91\%	&	13.51\%	&	27.06\%	\\ \hline \hline
R-EBP-P	&	2.78\%	&	2.29\%	&	2.28\%	&	2.20\%	&	2.25\%	&	2.16\%	&	2.22\%	&	2.22\%	&	2.10\%	\\ \hline
R-EBP-D	&	3.04\%	&	2.42\%	&	2.23\%	&	2.25\%	&	2.27\%	&	2.63\%	&	2.59\%	&	2.41\%	&	2.42\%	\\ \hline
\end{tabular}
 \label{tab:1d_without_dropout}
\end{table}

\textbf{Effects of Dropout} The results of EBP algorithms on different network structures with dropout are shown in Table~\ref{tab:1d_with_dropout}. The results show the same observations as those of without dropout. Comparing the results between Table~\ref{tab:1d_without_dropout} and Table~\ref{tab:1d_with_dropout}, we can see that using dropout can improve the performance in all configurations, which demonstrates that the dropout also works in the EBP algorithms. From the results of using 1000 units and 800  units in one-hidden-layer structure, we can see that without dropout, the result of 1000 hidden units is worse that that of 800 hidden units, while with dropout, the performance is continuously increasing when increase the hidden unit number from 800 to 1000. Besides, with dropout, the performance of B-EBP-D becomes reasonable. The results validate that dropout can effectively prevent overfitting in BMNNs with the EBP algorithm.
\begin{table}
\scriptsize
\centering
\caption{Test errors with dropout for 1D vector method}
\begin{tabular}{|c||c|c|c|c|c||c|c|c|c|} \hline
Hidden units	&	200	&	400	&	600	&	800	&	1000	&	[200, 200]	&	[400, 400]	&	[600, 600]	&	[800, 800]	\\ \hline \hline
B-EBP-P	&	3.60\%	&	2.82\%	&	2.82\%	&	2.55\%	&	2.52\%	&2.93\%	&2.39\%	&2.12\%	&2.12\%	\\ \hline
B-EBP-D	&	4.91\%	&	3.50\%	&	3.45\%	&	3.10\%	&	3.08\%	&3.97\%	&3.18\%	&2.89\%	&2.68\%	\\ \hline \hline
R-EBP-P	&	2.45\%	&	2.04\%	&	1.90\%	&	1.87\%	&	1.88\%	&2.22\%	&1.78\%	&1.75\%	&1.66\%	\\ \hline
R-EBP-D	&	2.58\%	&	2.09\%	&	1.94\%	&	1.91\%	&	1,86\%	&2.51\%	&1.99\%	&1.87\%	&1.75\%		\\ \hline
\end{tabular}

 \label{tab:1d_with_dropout}
\end{table}

\textbf{Effects of Spatial Filtering} Table~\ref{tab:2d_without_dropout} shows the results of MNNs using the EBP algorithm with the consideration of image spatial configuration. The best performance of spatial filtering method using binary weights is 3.56\% (obtained by 225 hidden units in one-layer structure), which is worse than the results of using ``1D Input Vector" method as shown in Table~\ref{tab:1d_without_dropout}. On the contrary, the performances of using real weights can be improved by the spatial filtering method, as the performance is better than all the network structures using ``1D Vector Input" method without dropout (the results in Table~\ref{tab:1d_without_dropout}). The best results are obtained in the configuration of 256 hidden units ($13 \times 13$ inputs to each unit in the hidden layer, and 256 inputs to each unit in the output layer). The results of this method shed light on the extension of the EBP method on Convolutional Neural Networks, such as the block size connecting to each unit in the feature map in each convolutional layer.
\begin{table}
\centering
\caption{Test errors without dropout for spatial filtering method}
\begin{tabular}{|c||c|c|c|c|c|c|c|c||c|} \hline
Hidden units	&	144	&	169	&	196	&	225	&	266	&	289	&	[361, 100]	\\ \hline \hline
B-EBP-P	&	4.06\%	&	3.90\%	&	3.87\%	&	3.97\%	&	4.07\%	&	4.36\%	&	4.96\%	\\ \hline
B-EBP-D	&	4.31\%	&	3.93\%	&	3.73\%	&	3.56\%	&	3.93\%	&	4.06\%	&	4.87\%	\\ \hline \hline
R-EBP-P	&	2.51\%	&	2.21\%	&	2.07\%	&	2.03\%	&	1.87\%	&	1.99\%	&	1.93\%	\\ \hline
R-EBP-D	&	2.82\%	&	2.51\%	&	2.18\%	&	2.22\%	&	2.17\%	&	2.08\%	&	2.02\%	\\ \hline

\end{tabular}
 \label{tab:2d_without_dropout}
\end{table}

\textbf{Summary} The analysis of experimental results gives us a few interesting findings. They include: (1) BMNNs with the EBP algorithm work well for image classification task, although the performance is not as good as real MNNs\footnote{Note that the EBP algorithm on MNNs with real weight can obtain comparable results with respect to the standard BackPropagation method.}; (2) even if the fan-in size is only few hundreds (e.g., [784, 200, 10]), the EBP algorithm still works well on BMNNs; (3) BMNNs with EBP-D algorithms on networks with two-hidden-layer (more layers) outperform the networks with one-hidden-layer; (4) dropout can significantly improve the performance of BMNNs with the EBP algorithm; and (5) BMNNs with the consideration of spatial filtering does not improve the classification performance, based on the results on MNIST.

\section{Conclusions}
In this paper, we report the performance of binary multilayer neural networks (BMNNs) on image classification tasks. Expectation BackPropagation (EBP) algorithm is used to train BMNNs with different network architectures and the performance is evaluated on the standard MNIST digits dataset. Experimental results demonstrate that BMNNs with the EBP algorithm can achieve good performance on the MNIST classification tasks. The results also show that the dropout techniques can significant improve BMNNs with the EBP algorithm. Image spatial configuration improves the performance of networks with real weights but not that of BMNNs. In this study, we only conduct experiments on the MNIST dataset. The performance of BMNNs with EBP algorithm on image classification tasks needs to be further validated on other image datasets (e.g., CIFAR10). In the future, we would like to study the performance of standard Convolutional Neural Networks with the use of EBP algorithm and to explore different weight initialization methods.

\subsubsection*{Acknowledgments}
 This is a course project for ``Lab Course on Deep Learning (11-875)", during Zhiyong Cheng's visit to Carnegie Mellon University.

\bibliographystyle{iclr2015}

\end{document}